\documentclass[conference]{IEEEtran}
\IEEEoverridecommandlockouts
\usepackage{cite}
\usepackage{amsmath,amssymb,amsfonts}
\usepackage{algorithmic}
\usepackage{graphicx}
\usepackage{textcomp}
\usepackage{xcolor}
\usepackage{multirow}
\usepackage{booktabs}
\usepackage{enumitem}
\usepackage{subfig}
\usepackage{float}
\usepackage{hyperref}
\def\BibTeX{{\rm B\kern-.05em{\sc i\kern-.025em b}\kern-.08em
    T\kern-.1667em\lower.7ex\hbox{E}\kern-.125emX}}

\def\vtp{p_\ell^{(i)}}
\def\vtm{m_\ell^{(i)}}
\def\hidden{\mathbf{h}_\ell^{(i)}}
\def\me{\mathbf{e}_\ell^{(i)}}
\def\ue{\mathbf{u}_\ell^{(i)}}
\def\ve{\mathbf{v}_\ell^{(i)}}
\def\temnei{\mathcal{N}_{{\rm Intra}(p_\ell^{(i)})}}
\def\socnei{\mathcal{N}_{{\rm Inter}(p_\ell^{(i)})}}
\def\temattn{\mathbf{t}_\ell^{(i)}}
\def\socattn{\mathbf{s}_\ell^{(i)}}

\def\hiddenmat{\mathcal{H}_{\mathcal{N}_{{\rm Intra}(p_\ell^{(i)})}}}
\def\temmat{\mathcal{T}_{\mathcal{N}_{{\rm Inter}(p_\ell^{(i)})}}}

\begin{document}

\title{Compatible Transformer for Irregularly Sampled \\ Multivariate Time Series

}

\author{%
\small
   Yuxi Wei$^{1*}$ \quad Juntong Peng$^1$ \quad Tong He$^2$ \quad Chenxin Xu$^1$ \quad Jian Zhang$^{2}$ \quad Shirui Pan$^{3}$ \quad Siheng Chen$^{1 \dag}$\\
$^1$ Cooperative Medianet Innovation Center, 
  Shanghai Jiao Tong University \\ $^2$ Amazon Web Service  \quad $^3$ University of Technology Sydney\\
 \texttt{\{wyx3590236732, juntong.peng, xcxwakaka, sihengc\}@sjtu.edu.cn}\\
\texttt{\{hetong007, zhjwyz\}@gmail.com} \quad
 \texttt{s.pan@griffith.edu.au}
\vspace{-5mm}
}

\maketitle

\renewcommand{\thefootnote}{\fnsymbol{footnote}}
\footnotetext[1]{Work done during internship at AWS. \qquad \footnotemark[2]Corresponding author.}

\begin{abstract}
To analyze multivariate time series, most previous methods assume regular subsampling of time series, where the interval between adjacent measurements and the number of samples remain unchanged. Practically, data collection systems could produce irregularly sampled time series due to sensor failures and interventions. However, existing methods designed for regularly sampled multivariate time series cannot directly handle irregularity owing to misalignment along both temporal and variate dimensions. To fill this gap, we propose Compatible Transformer (\texttt{CoFormer}), a transformer-based encoder to achieve comprehensive temporal-interaction feature learning for each individual sample in irregular multivariate time series. In~\texttt{CoFormer}, we view each sample as a unique variate-time point and leverage intra-variate/inter-variate attentions to learn sample-wise temporal/interaction features based on intra-variate/inter-variate neighbors. With \texttt{CoFormer} as the core, we can analyze irregularly sampled multivariate time series for many downstream tasks, including classification and prediction. 
 We conduct extensive experiments on 3 real-world datasets and validate that the proposed \texttt{CoFormer} significantly and consistently outperforms existing methods. Code will be avilable at \href{https://github.com/MediaBrain-SJTU/CoFormer}{https://github.com/MediaBrain-SJTU/CoFormer}.
\end{abstract}


\section{Introduction}
\label{sec:intro}
Multivariate time series is ubiquitous in many domains, such as business activities, industrial manufacturing, scientific experiments and multimedia~\cite{esmael2012multivariate}. It contains rich temporal information and stealthy interaction information among multiple variates. Multivariate time series analysis, including classification, regression and forecasting, is fundamental and pivotal in a wide range of real-world applications.

To effectively collect, process and store time series, modern sensors usually take samples regularly and uniformly in time to produce a sequence of numbers. To analyze those regularly sampled time series, researchers from diverse disciplines come up with numerous powerful models, such as Fourier transform, autoregressive model and Kalman filter. In the deep learning era, many recurrent neural networks, including LSTM and GRU, have achieved great success in temporal feature learning. To further capture correlations among multiple variates, graph neural networks~\cite{xu2022groupnet,xu2022dynamic} and attention operations~\cite{xu2023auxiliary} have also been adopted for interaction feature learning.
\begin{figure}[t]
  \centering
    \includegraphics[width=0.98\linewidth]{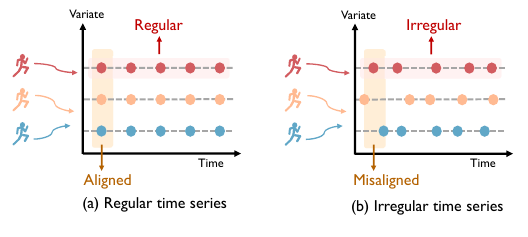}
    \vspace{-5mm}
   \caption{\small An illustration of regular and irregular multivariate time series. Irregularity causes misalignment along both time and variate dimensions. The proposed \texttt{CoFormer} handles irregular data.}
   \label{fig:introsketch}
    \vspace{-7mm}
\end{figure}

However, real-world data collection systems could produce irregularly sampled time series due to sensor failures, interventions or external forces in physical systems~\cite{zhang2021graph}. In these scenarios, multiple variates are associated with asynchronized, nonuniformly sampled measurements, which are not well aligned along either temporal or variate dimensions. Therefore, most models designed for traditional regularly sampled time series cannot be used to directly handle irregularly sampled multivariate time series. Imputation is an intuitive way to deal with irregularity, but it inevitably introduces noises or artifacts. Thus, some recent methods try to directly handle the irregularly sampled multivariate time series without imputation. Some previous works consider variate-centric modeling and separate temporal/interaction feature learning into two stages, which learn temporal features for each variate and then explore interactions between variates. This does not explicitly and comprehensively exploit the information in each single irregular sample.

In this work, we consider a novel sample-centric modeling method. The core idea is to view each irregular sample as an individual variate-time point, which makes each sample not rely on an aligned temporal or variate index, so that irregular sampling is no longer a matter. Based on this idea, we propose Compatible Transformer (\texttt{CoFormer}), a backbone encoder to learn comprehensive temporal and interaction information for each sample in irregularly sampled multivariate time series.  For each variate-time point, we define an intra-variate neighbor set and inter-variate neighbor set, each of which is a set of variate-time points, regardless of misalignment. Based on intra-variate/inter-variate neighbors, we propose point-set-based intra-variate/inter-variate attentions to extract temporal/interaction information. Note that the proposed \texttt{CoFormer} is well compatible with both regular and irregular scenarios since our intra-variate/inter-variate neighbors are also well-defined for irregularly sampled multivariate time series.

Based on the proposed \texttt{CoFormer}, we design systems to handle specific downstream tasks with irregularly sampled multivariate time series, respectively. We further conduct experiments on 3 real-world classification datasets, and \texttt{CoFormer} significantly outperforms state-of-the-art methods with AUROC improvement of 4.0\% and AUPRC improvement of 5.5\% on classification tasks. The method based on CoFormer can also be extended to related regression/prediction tasks.

The main contributions are summarized as follows:

    $\bullet$ We propose \texttt{CoFormer}, a transformer-based encoder to extract intra-variate temporal and inter-variate interaction information for each individual sample in irregularly sampled multivariate time series.
    
    $\bullet$ We propose \texttt{CoFormer}-based classification system for analyzing irregularly sampled multivariate time series.
    
    $\bullet$ We conduct experiments on 3 real-world datasets and show that i) the proposed \texttt{CoFormer} outperforms all the baselines significantly; ii) \texttt{CoFormer} shows outstanding flexibility and robustness.

\section{Related Works}
\label{sec:relatedworks}
\textbf{Multivariate time series analysis.}
Multivariate time series tasks have been studied for a long time. There are many downstream tasks related to multivariate time series such as forecasting, classification and anomaly detection. In these tasks, both temporal and interaction information are important, thus extracting information from both parts is one focus of the previous works. For temporal information, some works tackle the
forecasting problem by transforming the non-stationary process to stationary; recurrent neural
networks (RNNs) models are widely used to model the temporal dependencies. For interaction information, some works model the interaction information by social operators~\cite{xu2022remember}; graph structures are applied to model the interaction between variates~\cite{hu2020collaborative,xu2023eqmotion,tang2023collaborative}. Recently, attention mechanism and transformer structure show great power in sequential data, and are used for both temporal and interaction information extraction~\cite{mao2023leapfrog}. However, previous methods assume the sample rate and timestamps stay the same, therefore they cannot handle irregular sampled time series tasks in which the interval between adjacent samples are not fixed and the number of measurements of each variate need not to be the same.

\textbf{Irregular time series analysis.}
Irregular time series means that the timestamp of each measurement can be continuous but not fixed and discrete. This irregularity makes the data misaligned and many previous methods cannot be trivially modified to work. 
Some recent methods try to directly handle the irregularly sampled multivariate time series, and previous SOTA models mainly consider
two approaches: imputation-based methods and variate-centric modeling. 1)The imputation-based methods, such
as mTAND~\cite{shukla2021multi}, IP-Net~\cite{shukla2019interpolation} and $\text{DGM}^2$~\cite{wu2021dynamic}
, first convert irregular time
series to regular data or embeddings through imputation-based methods, and then apply
operations designed for the regular setting. But these converting processes bring inevitable noises and disturb the model learning. In comparison, our \texttt{CoFormer}
directly handles raw irregular data without approximation
errors during conversion. 2) The variate-centric modeling
methods, such as SeFT~\cite{horn2020set} and Raindrop~\cite{zhang2021graph}, combine all sample points of a variate together and then extract one unifying feature for each variate; that is, these methods aggregate
temporal information once, then followed by extracting interaction information at the level of variates once. In contrast, our \texttt{CoFormer} keep iteratively and alternatively extracting temporal and interaction information for each sample point. 


\begin{figure*}[t]
  \centering
    \includegraphics[width=0.98\linewidth]{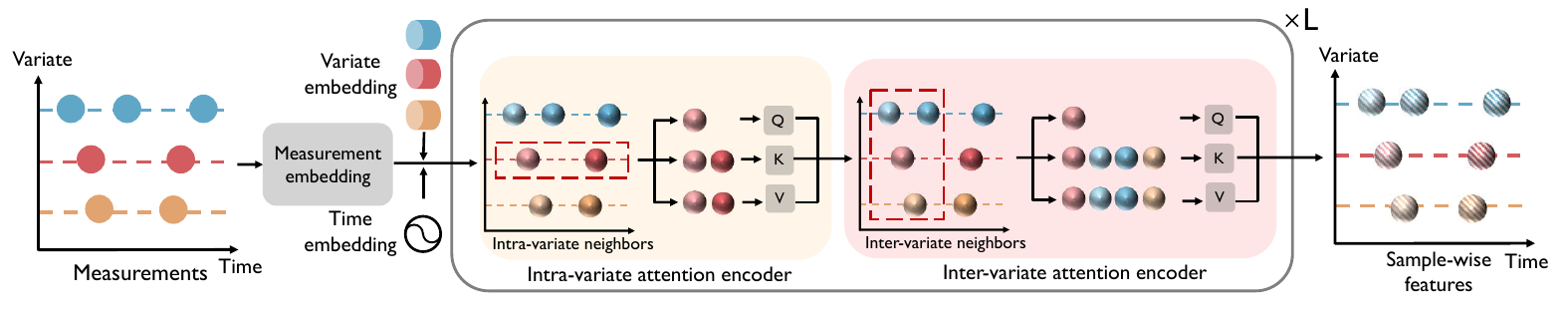}
    \vspace{-3mm}
   \caption{\small \textbf{The architechure of \texttt{CoFormer}}. \texttt{CoFormer} models each sample as variate-time point, constructs neighbors for the variate-time points, and applies intra-variate/inter-variate attention operation with the variate-time points and their corresponding neighbors.}
   \label{fig:encoder}
    \vspace{-5mm}
\end{figure*}

\section{Problem Formulation}
\label{sec:problem}
Mathematically, let $x$ be one continuous time series defined for all times $t \in \mathbb{R}$; that is, $x(t) \in \mathbb{R}$. In practice, we usually subsample the continuous time series $x$ to obtain a sequence of real numbers describing $x$. Let $\mathbf{m} = [m_1, m_2, \cdots, m_L]^T  \in \mathbb{R}^L$ be a sequence of $L$ historical samples, where $m_\ell = x(-t_\ell)$ is the measurement sampled at time stamp $t_\ell \geq 0$ from time series $x$. Given a set of $N$ sequences $\mathcal{M} = \{\mathbf{m}^{(1)}, \mathbf{m}^{(2)}, \cdots, \mathbf{m}^{(N)}\}$ measured from multivariate continuous time series $\mathcal{X} =\{ x^{(1)}, x^{(2)}, \cdots, x^{(N)}\}$, we aim to learn a model $f(\cdot)$, so that $\widehat{\mathcal{Y}} = f(\mathcal{M})$ is close to the ground-truth $\mathcal{Y}$. In the classification task, $\mathcal{Y}$ is the categories.

Many previous works consider regular subsampling, where each historical sample $\vtm{}$ is regularly measured from time series $x^{(i)}$ with the fixed time interval $\tau$; that is, the $\ell$-th measurement of the $i$-th variate is $\vtm{} = x^{(i)}(- \ell \cdot \tau)$. In this case, all sequences share the synchronized timestamps and  $\mathcal{M}$ can be well represented as a variate-time matrix. 

In this paper, we consider a general setting of irregular subsampling, where $\vtm{} = x^{(i)}(- t^{(i)}_{\ell})$, reflecting that each time interval for a sequence or a timestamp is unique and can be different. Moreover, each sequence $\mathbf{m}^{(i)}$ could have a different number of samples. Therefore, sequences in $\mathcal{M}$ are not well synchronized. It is then inappropriate to design the model $f(\cdot)$ by using standard tools designed based on a regularly sampled prior, such as convolution and recurrent operations. To handle irregularly sampled data, we face two challenges: i) how to extract temporally consistent features from each irregularly sampled sequence; and ii) how to socially fuse features with multiple irregularly sampled sequences.  The next section presents a Compatible Transformer, which leverages time encoding and novel attention designs to address the above two issues.

\section{Compatible Transformer}
\label{sec:method}

This section proposes Compatible Transformer (\texttt{CoFormer}), which effectively extracts temporal and interaction information from irregularly sampled multivariate time series. The core idea is to abstract each sample as a variate-time point, which has unique temporal and variate meaning without assuming regularity; and then, we leverage irregularity-compatible attention operations from temporal and variate aspects to exploit those variate-time points. 

Mathematically, let the variate-time point $\vtp{} = (i,t_\ell^{(i)})$ be the unique index of the $\ell$-th sample in the $i$-th variate, whose corresponding raw measurement is $\vtm{}$. The overall procedure of \texttt{CoFormer} is as follows,
\begin{subequations}
\setlength{\abovedisplayskip}{4pt}
   \setlength{\belowdisplayskip}{3pt}
\begin{eqnarray}
    \label{eq:measurement_encoding}
    \me{} & = & f_{\rm ME}(\vtm{}),
    \\
    \label{eq:position_encoding}
    \ue{}, \ve{} & = & f_{\rm VTE}(\vtp{}),
    \\
    \label{eq:aggregation}
    \hidden{} & = & f_{\rm AGG} (\me{}, \ue{}, \ve{}),
    \\
    \label{eq:temporal_attention}
    \temattn{} & = & f_{\rm Intra} (\hidden{}, \hiddenmat{}),
    \\
    \label{eq:interaction_attention}
    \socattn{} & = & f_{\rm Inter} (\temattn{}, \temmat{}).
\end{eqnarray}
\end{subequations}
Step~\eqref{eq:measurement_encoding} uses a measurement encoder $f_{\rm ME}(\cdot)$ to obtain the measurement embedding $\me{}$; Step~\eqref{eq:position_encoding} uses variate and time encoder $f_{\rm VTE}(\cdot)$ to obtain the variate code  $\ue{}$ and the time code $\ve{}$; Step~\eqref{eq:aggregation} aggregates the measurement embedding, variate code and  time code together to initialize the variate-time embedding $\hidden{}$; Step~\eqref{eq:temporal_attention} applies an intra-variate attention $f_{\rm Intra}(\cdot)$ to obtain the intra-variate-attentioned variate-time embedding $\temattn{}$, where $\temnei$ is a point set, defining the intra-variate neighbors of the point $\vtp{}$ and $\hiddenmat{}$ is the collection of intra-variate neighboring features; and Step~\eqref{eq:interaction_attention} applies an inter-variate attention $f_{\rm Inter}(\cdot)$ to obtain the inter-variate-attentioned variate-time embedding $\socattn{}$, where $\socnei$ is the inter-variate neighbors of the point $\vtp{}$ and $\temmat{}$ is the collection of  inter-variate neighbors' corresponding features after intra-variate attention.

Note that 
i) the proposed \texttt{CoFormer} realizes temporally consistent features in Steps~\eqref{eq:position_encoding} and~\eqref{eq:temporal_attention} as we encode the timestamp information in the feature embedding and leverage irregularity-compatible intra-variate attention to extract temporal information;
ii) the proposed \texttt{CoFormer} promotes comprehensive interaction features in Step~\eqref{eq:interaction_attention} as we apply inter-variate attention;
iii) the proposed \texttt{CoFormer} uses similar operations to handle both temporal and interaction information, which not only promotes the beauty of symmetry but also shows a general approach to get rid of regularity assumption, and compared with applying attention operation on all variate-time points, two kinds of neighbors make the information source more specific and reduce the time complexity from the square of all points to the square of the number of neighbors; and iv) Steps~\eqref{eq:temporal_attention} and~\eqref{eq:interaction_attention} can be applied successively for multiple times to obtain comprehensive temporal and interaction information. 

We now elaborate on the details of each step.

\textbf{Measurement embedding.} The functionality is to map a scalar measurement to high-dimensional embedding space for the subsequent process. Here we simply use a multilayer perceptron (MLP) to implement the measurement encoder $f_{\rm ME}(\cdot)$. The trainable parameters of the $\text{MLP}$ are shared across the measurements belonging to the same variate.

\textbf{Variate and time encoding.} Different from the position encoding in many other scenarios, here we have two dimensions and need to encode both variate and time.  To encode variate, we consider dictionary learning to convert the serial number of a variate $i$ into a variate code $\ve{} \in \mathbf{R}^d$~\cite{le2014distributed}. Each variate serial number $i$ is mapped into a trainable embedding vector, which corresponds to the $i$-th column of the dictionary. To encode time, we consider traditional trigonometric functions~\cite{vaswani2017attention}. Any continuous-valued time stamp $t_\ell^{(i)}$ can be mapped into a time code $\mathbf{u}_{\ell}^{(i)} \in \mathbf{R}^{\psi}$ through
\begin{equation}
    \setlength{\abovedisplayskip}{2pt}
   \setlength{\belowdisplayskip}{2pt}
  (\mathbf{u}_{\ell}^{(i)})_{2k} = \sin(\frac{t_\ell^{(i)}}{10000^{2k/ \psi}}),
  (\mathbf{u}_{\ell}^{(i)})_{2k+1} = \cos(\frac{t_\ell^{(i)}}{10000^{2k/ \psi}}),
  \label{eq:timeencoding}
\end{equation}
where $2k$ and $2k+1$ are the element indices of the code. Note that traditional position encoding encodes the fixed discrete positions, while we encode the continuous timestamp $t_\ell^{(i)}$ to handle irregular subsampling.

\textbf{Variate-time aggregation.} Here we attach the variate code and time code to the measurement embedding. The initial variate-time embedding is obtained as 
\begin{equation*}
    \setlength{\abovedisplayskip}{2pt}
   \setlength{\belowdisplayskip}{2pt}
    \hidden{} = \ve{} \oplus \text{Linear}(\me{} \oplus \ue{}),
\end{equation*}
where $\oplus$ is the concatenation operation and $\text{Linear}(\cdot)$ is a trainable linear layer. Because of aggregating variate and time codes, $\hidden{}$ inherits the context meaning and it works whatever the sample $\vtp{}$ is irregularly sampled or not.
 
\textbf{Intra-variate attention.} 
The functionality is to apply attention along the temporal dimension to learn long-range intra-variate temporal information. We construct intra-variate neighboring set for each variate-time point. Let $\temnei{} = \{p_1^{(i)},p_2^{(i)},\cdots,p_L^{(i)}\}$ be a point set that contains all the intra-variate neighbors of $\vtp{}$; that is, all the variate-time points belonging to the same $i$-th variate. We next collect the corresponding features from each point in $\temnei{}$; that is, $\hiddenmat{}= \{ \mathbf{h}_1^{(i)},\mathbf{h}_2^{(i)},\cdots,\mathbf{h}_L^{(i)}\}$. Finally, we leverage the standard multi-head attention to learn intra-variate temporal information. The intra-variate-attentioned embedding of $\vtp{}$ is obtained as,
\begin{equation*}
    \setlength{\abovedisplayskip}{2pt}
   \setlength{\belowdisplayskip}{2pt}
    \temattn{} = \text{MHA}_{\rm Intra}(\hidden{},\hiddenmat{}{},\hiddenmat{}),
\end{equation*}
where $\text{MHA}_{\rm Intra}(\cdot)$ denotes the multi-head attention among the intra-variate neighbors. Since attention operation can handle any length of inputs, our intra-variate attention encoder can deal with sequences with arbitrary length and any irregularly sampled timestamp.

\textbf{Inter-variate attention.}
Multiple time series might contain rich correlated information. To exploit interactions, we apply attention within the features belonging to different variates to comprehensively learn  short-range interaction information. In most scenarios, interactions are captured from multiple variates at the same timestamps. Since we deal with irregular sampling, variates are asynchronized and we thus need to consider more general inter-variate neighbors. Let $\socnei{} = \{p_m^{(j)}||t_m^{(j)}-t_\ell^{(i)}| \leq \Delta \tau\}$ be a point set for inter-variate neighbors of $\vtp{}$, which includes all the variate-time points whose sampled timestamp is close to $t_\ell^{(i)}$, and the  inter-variate interaction information is extracted by multi-head attention among them. The intuition is that the smaller the interval between two timestamps is, the closer this relation between the samples is. For instance, the players usually predict the intention of others by the current positions, and medical sensors' measurements would fluctuate in a period together due to seizure of diseases. Note that i) the definition of $\socnei{}$  is compatible for both regular and irregular cases as when $\Delta \tau = 0$, $\socnei{}$ degenerates to the common synchronized setting; ii) instead of assigning a radius, we could also consider $K$-nearest neighbors in the temporal dimension, which is equivalent to some extend. Without annotation, the experiments follow the $K$-nearest neighbors implementation.

Given the inter-variate neighborhood, the rest procedure is similar to intra-variate attention. Let $\temmat{}=\{\mathbf{t}_m^{(j)}|p_m^{(j)} \in \socnei{}\}$ be the intra-variate-attentioned features collected from all the points in $\socnei{}$. We then obtain the inter-variate-attentioned embedding as
\begin{equation*}
\setlength{\abovedisplayskip}{2pt}
   \setlength{\belowdisplayskip}{2pt}
    \socattn{} = \text{MHA}_{\rm Inter}(\temattn{},\temmat{},\temmat{}),
\end{equation*}
where $\text{MHA}_{\rm Inter}(\cdot)$ denotes the multi-head attention among the inter-variate neighbors. 

\textbf{Successive feature learning.}
To comprehensively learn temporal and interaction information, we successively apply intra-variate/inter-variate attention for multiple times. A complete successive attention layer consists of intra-variate attention, layernorm~\cite{ba2016layer}, residual connection~\cite{he2016deep} and  inter-variate attention. We stack multiple successive attention layers to enrich the representation capability.

\section{Classification System}
\vspace{-1mm}
\begin{figure}

    \centering
    \includegraphics[width=0.95\linewidth]{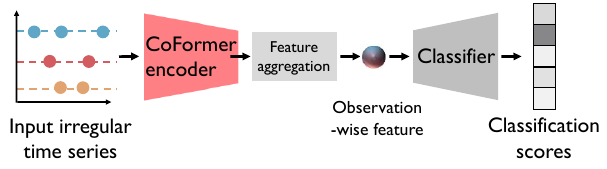}
    \vspace{-2mm}
    \caption{\small Classification system of \texttt{CoFormer}. \texttt{CoFormer} extracts features from irregular input directly. The extracted sample-wise features are aggregated as the final observation-wise feature and the classification score is output by a classifier. }
    \label{fig:classification}
    \vspace{-5mm}
\end{figure}

Consider the entire irregularly sampled multivariate time series as one observation. Here the specific aim is to classify an input observation into a predefined category. Our classification system consists of three modules: the proposed \texttt{CoFormer} for sample-wise feature learning, the aggregation module for observation-wise feature aggregation, and the classification module for discrimination; see Figure~\ref{fig:classification}.

\textbf{Sample-wise feature learning.} We use the proposed \texttt{CoFormer} to learn features for each irregular sample. Let $\mathcal{M}$ be an input observation of irregularly sampled multivariate time series. The collection of sample-wise features is obtained as
$
    \mathcal{F} = \{ \mathbf{f}_\ell^{(i)} \}_{i, \ell} = \text{\texttt{CoFormer}}(\mathcal{M}),
$
where $\mathbf{f}_\ell^{(i)}$ is the feature embedding of the sample $p_\ell^{(i)}$ output by the last inter-variate attention. Note that $i$ is the variate index, ranging from $1$ to $N$; while $\ell$ is the time index and each variate might have a different number of samples.

\textbf{Observation-wise feature aggregation.} To realize classification, we need to aggregate all the sample-wise features to be a unified feature for the entire observation. To aggregate along the temporal dimension and get the variate-wise feature, we first take the average: $\textbf{f}_{\rm mean}^{(i)} = \sum_{\ell=1}^L \mathbf{f}_\ell^{(i)}/L$, where $L$ is the sample number of the $i$-th variate. Since averaging does not take each sample's importance into account, we further use an attention operator to introduce importance; that is,
$
    \textbf{f}^{(i)} = \text{MHA}( \textbf{f}_{\rm mean}^{(i)}, \mathcal{F}^{(i)}, \mathcal{F}^{(i)}),
$
where $\text{MHA}(\cdot)$ denotes the multi-head attention and $\mathcal{F}^{(i)} = \{ \mathbf{f}_\ell^{(i)} \}_{\ell=1}^L$ is a collection of the $i$-th variate's feature. To aggregate along the variate dimension and get the observation-wise feature, we simply take the average and the final observation-wise feature is $\textbf{f} = \sum_{i=1}^N \textbf{f}^{(i)}/N$.

\textbf{Classification.} The observation-wise feature is input into a common readout layer classifier implemented by MLP which outputs the final classification score.

\textbf{Loss function.}
The common loss functions designed for classification task can be used. Here we consider the cross-entropy loss in our experiments.

\section{Experiments}
\label{sec:experiments}

\begin{table*}
    \centering
    \small
    \vspace{-1mm}
    \caption{\small Classification results of irregular methods and \texttt{CoFormer}. \texttt{CoFormer} is the best model on the 3 irregular multivariate classification datasets. We use 5 different splits of the  datasets and calculate the mean and variance of the 5 results. The splits are fixed across all methods. Imputat. indicates imputation-based methods, and Var. indicates variate-centric methods. }
    \vspace{-1mm}
    \resizebox{150mm}{27mm}{
    \begin{tabular}{c c  c c c c  c c c c}
    \cmidrule(r){2-10}
          ~ &\multirow{2}*{Method} & \multicolumn{2}{c}{P19} & \multicolumn{2}{c}{P12} & \multicolumn{4}{c}{PAM} \\ 
         \cmidrule(r){3-4}  \cmidrule(r){3-4}  \cmidrule(r){5-6}  \cmidrule(r){7-10}  
         ~ & ~  & AUROC  &  AUPRC  & AUROC  &  AUPRC  &  Accuracy  &  Precision  &  Recall  &  F1 score\\
         \cmidrule(r){2-2} \cmidrule(r){3-4}  \cmidrule(r){3-4}  \cmidrule(r){5-6}  \cmidrule(r){7-10} 
          ~ & Transformer~\cite{vaswani2017attention}  &  83.2$\pm$1.3 &  47.6$\pm$3.8 & 65.2$\pm$5.6  &  95.7$\pm$1.6  &  83.5$\pm$1.5  &  84.8$\pm$1.5  &  86.0$\pm$1.2  &  85.0$\pm$1.3 \\
          ~ &Transformer-mean   &  84.1$\pm$1.7 &  47.4$\pm$1.4 & 66.8$\pm$4.2  &  95.9$\pm$1.1  &  83.7$\pm$2.3  &  84.9$\pm$2.6  &  86.4$\pm$2.1  &  85.1$\pm$2.4 \\
          ~ &MTGNN ~\cite{wu2020connecting}    & 81.9$\pm$6.2 & 39.9$\pm$8.9 & 67.5$\pm$3.1 & 96.4$\pm$0.7 & 83.4$\pm$1.9 & 85.2$\pm$1.7 & 86.1$\pm$1.9 & 85.9$\pm$2.4 \\
          ~ &GRU-D~\cite{che2018recurrent}   &  83.9$\pm$1.7 &  46.9$\pm$2.1 & 67.2$\pm$3.6  &  95.9$\pm$2.1  &  83.3$\pm$1.6  &  84.6$\pm$1.2  &  85.2$\pm$1.6  &  84.8$\pm$1.2 \\
          ~ &stop\&hop~\cite{hartvigsen2022stop}  & 81.5$\pm$2.3 & 47.0$\pm$3.1 & 68.5$\pm$2.2 & 96.2$\pm$0.8 & 84.1$\pm$1.7 & 84.9$\pm$1.5 & 86.2$\pm$2.0 & 85.5$\pm$1.7 \\
         \cmidrule(r){2-10}
         \multirow{3}*{\rotatebox{90}{ Imputat.}} & 
         mTAND~\cite{shukla2021multi}   &  80.4$\pm$1.3 &  32.4$\pm$1.8 & 65.3$\pm$1.7  &  96.5$\pm$1.2  &  74.6$\pm$4.3  &  74.3$\pm$4.0  &  79.5$\pm$2.8  &  76.8$\pm$3.4 \\
          ~ &IP-Net~\cite{shukla2019interpolation}  &  84.6$\pm$1.3 &  38.1$\pm$3.7 &72.5$\pm$2.4  &  96.7$\pm$0.3  &  74.3$\pm$3.8  &  75.6$\pm$2.1  &  77.9$\pm$2.2  &  76.6$\pm$2.8 \\
          ~ & $\text{DGM}^2-\text{O}$~\cite{wu2021dynamic}   &  86.7$\pm$3.4 &  44.7$\pm$11.7 &71.2$\pm$2.5  &  96.9$\pm$0.4  &  82.4$\pm$2.3  &  85.2$\pm$1.2  &  83.9$\pm$2.3  &  84.3$\pm$1.8 \\
         \cmidrule(r){2-10}
         \multirow{2}*{\rotatebox{90}{ Var.}} & SeFT~\cite{horn2020set}   &  78.7$\pm$2.4  &31.1$\pm$2.8 & 66.8$\pm$0.8  &  96.2$\pm$0.2  &  67.1$\pm$2.2  &  70.0$\pm$2.4  &  68.2$\pm$1.5  &  68.5$\pm$1.8 \\
          ~ & Raindrop~\cite{zhang2021graph}   &  87.0$\pm$2.3 &  51.8$\pm$5.5 &72.1$\pm$1.3  &  97.0$\pm$0.4  &  88.5$\pm$1.5  &  89.9$\pm$1.5  &  89.9$\pm$0.6  &  89.8$\pm$1.0 \\
         \cmidrule(r){2-2} \cmidrule(r){3-4}  \cmidrule(r){3-4}  \cmidrule(r){5-6}  \cmidrule(r){7-10} 
          ~ &\textbf{CoFormer} (Ours) & \textbf{89.2 }$\pm$\textbf{1.8} & \textbf{57.3}$\pm$\textbf{3.3} & \textbf{76.5}$\pm$\textbf{2.9} & \textbf{97.5}$\pm$\textbf{0.5} & \textbf{91.2}$\pm$\textbf{0.6} & \textbf{92.4}$\pm$\textbf{0.7} & \textbf{93.7}$\pm$\textbf{0.7 } & \textbf{92.8}$\pm$\textbf{0.5}\\
         \cmidrule(r){2-10}
    \end{tabular}
    }
    \label{tab:classification}
    \vspace{-3mm}
\end{table*}

\subsection{Datasets and metrics}
We evaluate our model on three healthcare and human activity classification datasets:(1) \textbf{P19}:P19~\cite{reyna2019early}(PhysioNet Sepsis Early Prediction Challenge 2019) dataset contains 38,803 patients, where each patient is monitored by 34 irregularly sampled sensors. We follow the pre-processes in \cite{zhang2021graph}. Each patient is labelled by occurrence of sepsis within the next 6 hours. (2) \textbf{P12}:P12~\cite{goldberger2000physiobank}(PhysioNet Mortality Prediction Challenge 2012) dataset includes 11,988 patients (observations), after removing 12 inappropriate observations following~\cite{zhang2021graph}. Each observation contains multivariate time series with 36 sensors. Each observation is associated with a binary label which indicates the length of stay in ICU, where negative label means hospitalization is not longer than 3 days and positive label marks hospitalization is longer than 3 days.(3) \textbf{PAM}:PAM~\cite{reiss2012introducing} (PAMAP2 Physical Activity Monitoring) dataset measures daily living activities of 9 subjects with 3 inertial measurement units. Following~\cite{zhang2021graph}, we modify it to suit the scenario of irregular time series classification. After modification, PAM dataset contains 5,333 segments (observations) of sensory signals. Each observation is measured by 17 sensors and contains 600 continuous samples with a sampling frequency 100 Hz. To make the time series irregular, we randomly remove 60\% samples. PAM is labelled into 8 classes where each class represents an activity of daily living.

Following~\cite{zhang2021graph}, \textbf{AUROC} and \textbf{AUPRC} are used as metrics on \textbf{P19} and \textbf{P12} datasets; \textbf{Accuracy}, \textbf{Precision}, \textbf{Recall} and \textbf{F1} score are used on $\textbf{PAM}$ dataset. Following ~\cite{alahi2016social}.

\subsection{Implement details}

In classification tasks, following settings in~\cite{zhang2021graph}, we use a batch size of 128, a learning rate of 0.0001 and an epoch of 20 in 3 classification datasets. The other parameters of the baselines are consistent with~\cite{zhang2021graph}, too. We use Adam optimizer for gradient-based optimization.
For \texttt{CoFormer}, all of the MLPs are 3 layers with ReLU as an activation function. All of the hidden sizes are set as 256, including the time code. The variate code dimension is 32. The number of multi-heads is 8 for all attention operations. The batch size, learning rate, epoch and optimizer are the same as the baselines. Note that, $\textbf{P19}$ and $\textbf{P12}$ datasets contain static vectors for each observation. Following~\cite{zhang2021graph}, we use a linear projection to map it into a high dimension and concatenate it with the observation-wise feature before the classifier. The number of successive layers is 4, the number of neighbors is 30.

\subsection{Quantitative results on classification tasks}


We randomly spilt the datasets into training (80\%), validation(10\%) and test(10\%) set following~\cite{zhang2021graph}. The indices of these splits are fixed across all methods. Table~\ref{tab:classification} compares \texttt{CoFormer} with a series of state-of-the-art time series classification methods.  The original MTGNN~\cite{wu2020connecting} cannot handle irregular data. To adapt it to our setting, we impute the missing data with linear interpolation for preprocessing. To make Transformer~\cite{vaswani2017attention} deal with common regular data, Transformer-mean also uses linear interpolation; see more details in the supplementary material. The experimental results show that our \texttt{CoFormer} gets the best performance across all of the datasets and all the metrics.  Compared to Raindrop\cite{zhang2021graph}, \texttt{CoFormer} outperforms with average AUROC improvement of
4.0\% and average AUPRC improvement of 5.5\% on $\textbf{P19}$ and $\textbf{P12}$ datasets. \texttt{CoFormer} shows strong feature learning capability for irregular multivariate time series. 

\vspace{-1mm}
\vspace{-1mm}
\subsection{Ablation studies}
\vspace{-1mm}

\begin{table}
    \centering
    \caption{\small Ablation study of each component, including intra-variate attention encoder (Intra-A), inter-variate attention encoder (Inter-variate A), time encoding (TE), variate encoding (VE) and attention aggregator (AR) on $\textbf{P19}$ . Each component shows benefits to the model.  
    }

    \small
    \resizebox{75mm}{12mm}{
    \begin{tabular}{c c c c c  c c}
    \toprule
         Intra-A & Inter-A & TE & VE & AR & AUROC & AUPRC\\
        \cmidrule(r){1-5} \cmidrule(r){6-7}
          \checkmark& \checkmark & \quad & \quad & \quad & 82.1$\pm$1.4 &47.3$\pm$1.7 \\
          \checkmark& \checkmark & \checkmark & \quad & \quad & 84.5$\pm$2.1 &49.6$\pm$2.2 \\
          \checkmark& \checkmark & \checkmark & \checkmark & \quad & 85.1$\pm$2.4 &50.2$\pm$2.9 \\
          \checkmark& \quad & \checkmark & \checkmark & \checkmark & 85.3$\pm$1.9&51.7$\pm$2.6 \\
          \quad& \checkmark & \checkmark & \checkmark & \checkmark & 86.6$\pm$2.2 &53.2$\pm$2.5 \\
          \checkmark& \checkmark & \checkmark & \checkmark & \checkmark & \textbf{89.2}$\pm$\textbf{1.8} &\textbf{57.3}$\pm$\textbf{3.3} \\
           
    \bottomrule
    \end{tabular}
    }
    \vspace{-4mm}
    \label{tab:components}
\end{table}

\textbf{Effect of architecture components.}
We validate the influence of each component introduced in Section~\ref{sec:method}, including intra-variate attention encoder (Intra-A), inter-variate attention encoder (Inter-A), time encoding (TE), variate encoding (VE) and attention aggregator (AR), as shown in Table~\ref{tab:components}. The full \texttt{CoFormer} achieves the best performance, indicating that every component is useful to the model.

\begin{figure}
    \centering
    \vspace{-1mm}
\includegraphics[width=0.92\linewidth]{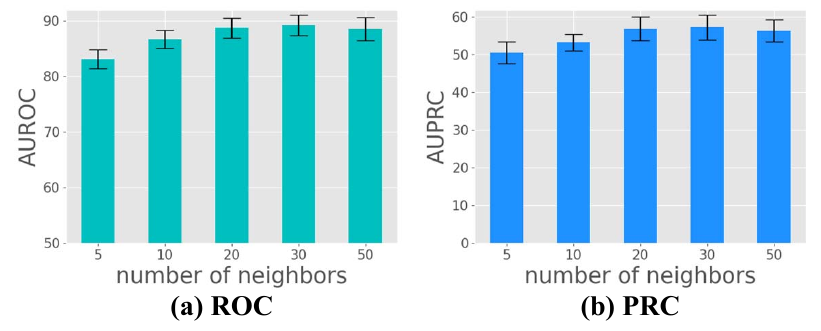}
    \vspace{-2mm}
    \caption{\small Ablation study of number of neighbors $K$ on $\textbf{P19}$ dataset. As $K$ increases, the performance also increases in the beginning, while a too large $K$ slightly worsens it.}
    \label{fig:neighbors}
    \vspace{-3mm}
\end{figure}

\textbf{Effect of the number of $K$ nearest neighbors.}
The effect of the number of the $K$ nearest neighbors is shown in Figure~\ref{fig:neighbors}.  As the $K$ increases, the performance 
also increases at the beginning, but a too large $K$ slightly worsen it. Based on the figure we suggest setting $K$ to be close to the number of variates, as in this way the inter-variate neighbors tend to cover samples from different and most of the variates. Besides, the performance only changes slightly with a large range of different values of $K$, which suggests that \texttt{CoFormer} is not sensitive to the tuning of this parameter.

\begin{figure}
    \vspace{-1mm}
    \centering
    \includegraphics[width=0.92\linewidth]{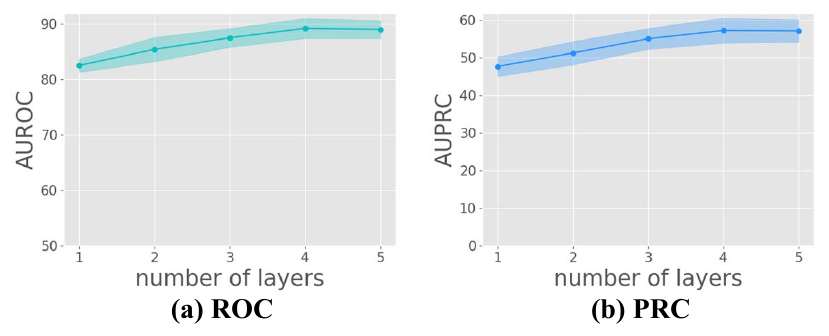}
    \vspace{-2mm}
    \caption{\small Ablation study of number of layers on $\textbf{P19}$ dataset. The performance increases first, and then becomes stable. }
    \vspace{-5mm}
    \label{fig:layers}

\end{figure}
\textbf{Effect of the number of successive layers.}
We explore the effect of the number of successive attention layers and illustrate in Figure~\ref{fig:layers}. Conceptually, a larger number of successive layers brings stronger representation capability,
and that matches the results from 1 to 4 layers. With 4 layers
the model complexity saturates for \textbf{P19} dataset, thus increasing
it further does not provide further performance improvements.

\vspace{-1mm}
\section{Conclusion}
\vspace{-1mm}
\label{sec:conclusion}
We propose \texttt{CoFormer}, a transformer-based encoder that works for both regularly and irregularly sampled multivarite time series. \texttt{CoFormer} views each sample as a variate-time point, and applies intra-variate/inter-variate attention operation to extract both temporal/interaction information from corresponding neighbors. Experiments reveal that \texttt{CoFormer} outperforms existing baselines and shows outstanding flexibility and robustness under various irregular settings.

\section*{ACKNOWLEDGEMENT}
This research is supported by NSFC under
Grant 62171276 and the Science and Technology Commission of Shanghai Municipal under Grant 21511100900 and
22DZ2229005.

\bibliographystyle{./CoS-TR/IEEEtran}
\bibliography{egbib}

\end{document}